%% file: main.tex
\documentclass[10pt,twocolumn,letterpaper]{article}

\usepackage[pagenumbers]{cvpr} 

\input{preamble}

\definecolor{cvprblue}{rgb}{0.21,0.49,0.74}
\usepackage[pagebackref,breaklinks,colorlinks,citecolor=cvprblue]{hyperref}

\usepackage{bm}
\usepackage{multirow}
\usepackage{colortbl}
\usepackage{comment}
\usepackage[accsupp]{axessibility} 

\newcommand{\ts}{\textsuperscript}

\definecolor{green}{HTML}{E2EFDA}
\definecolor{yellow}{HTML}{FFF2CC}

\def\paperID{3201} 
\def\confName{CVPR}
\def\confYear{2024}

\newcommand{\OURS}{GaussianAvatars}
\title{\OURS: Photorealistic Head Avatars with Rigged 3D Gaussians}

\author{
Shenhan Qian$^{1}$~~~~~~~~
Tobias Kirschstein$^1$~~~~~~~~
Liam Schoneveld$^2$~~~~~~~~
Davide Davoli$^{3}$\\
Simon Giebenhain$^1$~~~~~~~~
Matthias Nie{\ss}ner$^1$
\vspace{0.2cm} \\
{
\setlength{\tabcolsep}{15pt} 
\renewcommand{\arraystretch}{0.2} 
\begin{tabular}[t]{c c c}%
  $^1$Technical University of Munich &
    $^2$Woven by Toyota &
    $^3$Toyota Motor Europe NV/SA\\ & & 
    \footnotesize{associated partner by contracted service}
\end{tabular}
}
}

\begin{document}

\twocolumn[{%
\renewcommand\twocolumn[1][]{#1}%
\maketitle
\vspace{-0.5cm}
\begin{center}
    \centering
    \captionsetup{type=figure}
    \includegraphics[width=1\linewidth ]{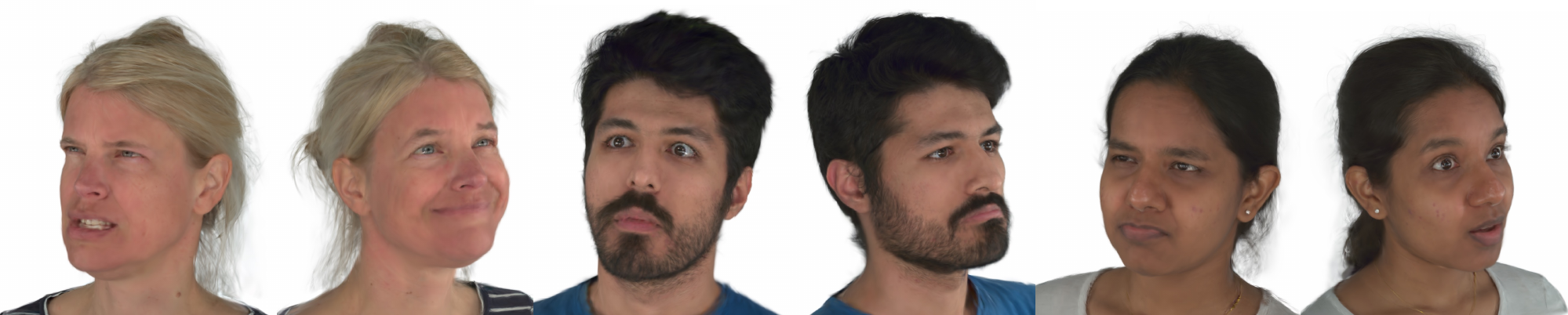}
    \caption{We introduce \OURS, a new method to create photorealistic head avatars from multi-view videos. Our avatar is represented by 3D Gaussian splats rigged to a parametric face model. We can fully control and animate our avatars in terms of pose, expression, and view point as shown in the example renderings above.}
\end{center}%

}]
\maketitle

\input{sec/0_abstract}
\input{sec/1_introduction}

\input{sec/2_related_work}

\input{sec/3_method}
\input{sec/4_experiments}

\input{sec/5_conclusion}

\input{sec/x_suppl}

{
    \small
    \bibliographystyle{ieeenat_fullname}
    \bibliography{main}
}


\end{document}

%% file: preamble.tex
\usepackage[dvipsnames]{xcolor}


%% file: sec/0_abstract.tex
\begin{abstract}
We introduce \OURS{}\footnote{Project page: \href{https://shenhanqian.github.io/gaussian-avatars}{https://shenhanqian.github.io/gaussian-avatars}}, a new method to create photorealistic head avatars that are fully controllable in terms of expression, pose, and viewpoint. The core idea is a dynamic 3D representation based on 3D Gaussian splats that are rigged to a parametric morphable face model. This combination facilitates photorealistic rendering while allowing for precise animation control via the underlying parametric model, e.g., through expression transfer from a driving sequence or by manually changing the morphable model parameters. We parameterize each splat by a local coordinate frame of a triangle and optimize for explicit displacement offset to obtain a more accurate geometric representation. During avatar reconstruction, we jointly optimize for the morphable model parameters and Gaussian splat parameters in an end-to-end fashion. We demonstrate the animation capabilities of our photorealistic avatar in several challenging scenarios. For instance, we show reenactments from a driving video, where our method outperforms existing works by a significant margin.
\end{abstract}

%% file: sec/1_introduction.tex
\section{Introduction}
\label{sec:intro}

Creating animatable avatars of human heads has been a longstanding problem in computer vision and graphics. In particular, the ability to render photorealistic dynamic avatars from arbitrary viewpoints enables numerous applications in gaming, movie production, immersive telepresence, and augmented or virtual reality. For such applications, it is also crucial to be able to control the avatar, and for it to generalize well to novel poses and expressions.

Reconstructing a 3D representation able to jointly capture the appearance, geometry and dynamics of human heads represents a major challenge for high-fidelity avatar generation.
The under-constrained nature of this reconstruction problem significantly complicates the task of achieving a representation that combines novel-view rendering photorealism with expression controllability.
Moreover, extreme expressions and facial details, like wrinkles, the mouth interior, and hair, are difficult to capture, and can produce visual artifacts easily noticed by humans.

Neural Radiance Fields (NeRF \cite{mildenhall2021nerf}) and its variants \cite{barron2021mip, chen2022tensorf, muller2022instant} have shown impressive results in reconstructing static scenes from multi-view observations.
Follow-up works have extended NeRF to model dynamic scenes for both arbitrary \cite{park2021nerfies, park2021hypernerf, li2022neural} and human-tailored scenarios \cite{kirschstein2023nersemble, isik2023humanrf}. These works achieve impressive results for novel view rendering; however, they lack controllability and as such do not generalize well to novel poses and expressions.

The recent 3D Gaussian Splatting method \cite{kerbl20233d} achieves even higher rendering quality than NeRF for novel-view synthesis with real-time performance, optimizing for discrete geometric primitives (3D Gaussians) throughout 3D space. 
This method has been extended to capture dynamic scenes by building explicit correspondences across time steps \cite{luiten2023dynamic}; however, they do not allow for animations of the reconstructed outputs.

To this end, we propose \OURS{}, a method for dynamic 3D representation of human head based on 3D Gaussian splats that are rigged to a parametric morphable face model.
Given a FLAME \cite{FLAME:SiggraphAsia2017} mesh, we initialize a 3D Gaussian at the center of each triangle. When the FLAME mesh is animated, each Gaussian then translates, rotates, and scales according to its parent triangle. 
The 3D Gaussians then form a radiance field on top of the mesh, compensating for regions where the mesh is either not accurately aligned, or is incapable of reproducing certain visual elements. To achieve high fidelity of the reconstructed avatar, we introduce a binding inheritance strategy to support  Gaussian splats without losing controllability. We also explore balancing fidelity and robustness to animate the avatars with novel expressions and poses.
Our method outperforms existing works by a significant margin on both novel-view rendering and reenactment from a driving video.

\smallskip
\noindent Our contributions are as follows:
\begin{itemize}
    \item We propose \OURS, a method to create animatable head avatars by rigging 3D Gaussians to parametric mesh models.
    \item We design a binding inheritance strategy to support adding and removing 3D Gaussians without losing controllability.
\end{itemize}

%% file: sec/2_related_work.tex
\section{Related Work}
\label{sec:related}

\subsection{Radiance Field Reconstruction}
NeRF \cite{mildenhall2021nerf} stores the radiance field of a scene in a neural network and provides photorealistic renderings of novel views with volumetric rendering. Later work \cite{yu2021plenoxels, sun2022direct} represents the scene as voxel grids, achieving comparable rendering quality in a shorter time. Efficiency can be further improved by employing compression methods such as voxel hashing \cite{muller2022instant} or tensor decomposition \cite{chen2022tensorf}. PointNeRF \cite{xu2022point} uses point clouds as a scene representation, whereas 3D Gaussian Splatting \cite{kerbl20233d} uses anisotropic 3D Gaussians that are rendered by sorting and rasterization, achieving superior visual quality with real-time rendering. Mixture of Volumetric Primitives~\cite{lombardi2021mixture} uses surface-aligned volumes to achieve fast rendering with high visual fidelity. It also applies residual transformations to primitives.
Our method follows 3D Gaussian Splatting \cite{kerbl20233d}, benefiting from the expressiveness of anisotropic Gaussians. 

A common paradigm to model a dynamic scene is directly storing it in an MLP with 4D coordinates as the input \cite{gao2021dynamic, xian2021space} or a 4D tensor with the time dimension or spatial dimensions compressed \cite{song2023nerfplayer, fridovich2023k, cao2023hexplane, attal2023hyperreel}. These methods can faithfully replay a dynamic scene and produce realistic novel-view rendering but lack an explicit handle to manipulate the content.
Another paradigm learns a static canonical space and the maps time steps back to the canonical space via separate MLPs \cite{pumarola2021d, park2021nerfies, park2021hypernerf}. 
Instead of using a deformation MLP, a proxy geometry provides more direct controllability. \citet{liu2021neural} warp points from an observed space to the canonical space based on the movement of the nearest triangle on an SMPL \cite{loper2023smpl} mesh. \citet{peng2021animatable} deform points with the skeleton of SMPL and neural blending weights. 
Concurrent works create human body avatars with forward deformation \cite{li2023animatablegaussians, lei2023gart, ye2023animatable, jena2023splatarmor, kocabas2023hugs} and cage-based deformation \cite{Zielonka2023Drivable3D}. Unlike these methods, we directly attach 3D Gaussians to triangles and explicitly move them, obviating the need for a canonical space and enabling effective mesh finetuning.

\subsection{Human Head Reconstruction and Animation}
Head avatar creation advanced through the uptake of differentiable rendering and scene representations.
\citet{thies2016face2face} instigated a shift toward digital avatars with real-time face tracking and authentic face reenactment.
Advancements in image synthesis with neural networks \cite{thies2019deferred, chan2019everybody} boosted the controllability of head avatars from lip-syncing to expression transfer and head motions~\cite{suwajanakorn2017synthesizing,kim2018deep,xu2023latentavatar}. \citet{gafni2021dynamic} learn a NeRF conditioned on an expression vector from monocular videos.
\citet{grassal2022neural} subdivide and add offsets to FLAME \cite{FLAME:SiggraphAsia2017} to enhance its geometry and enable a dynamic texture via an expression-dependent texture field.
IMavatar~\cite{zheng2022avatar} learns a 3D morphable head avatar with neural implicit functions, solving for a map from observed to canonical space via iterative root-finding.
HeadNeRF \cite{hong2022headnerf} learns a NeRF-based parametric head model with 2D neural rendering for efficiency.

INSTA~\cite{zielonka2023instant} deforms query points to a canonical space by finding the nearest triangle on a FLAME~\cite{FLAME:SiggraphAsia2017} mesh, and combines this with InstantNGP~\cite{muller2022instant} to achieve fast rendering.
Like INSTA, we also use triangles to warp the scene, but we build a consistent correspondence between a 3D Gaussian and a triangle instead of querying the nearest one for each timestep. 
\citet{zheng2023pointavatar} explore point-based representations with differential point splatting. They define a point set in canonical space and learn a deformation field conditioned on FLAME's expression vectors to animate the head avatar. While the scale of a point has to be manually set for their method, it is an optimizable parameter for 3D Gaussians. 
NeRFBlendShape \cite{Gao2022nerfblendshape} models a dynamic scene by blending hash tables with 3DMM parameters \cite{blanz2003face, paysan20093d}. 
AvatarMAV \cite{xu2023avatarmav} decouples motion and appearance, blending voxel grids only for the motion field.

%% file: sec/3_method.tex
\section{Method}

\begin{figure*}[h]
  \centering
  \includegraphics[width=1\textwidth ]{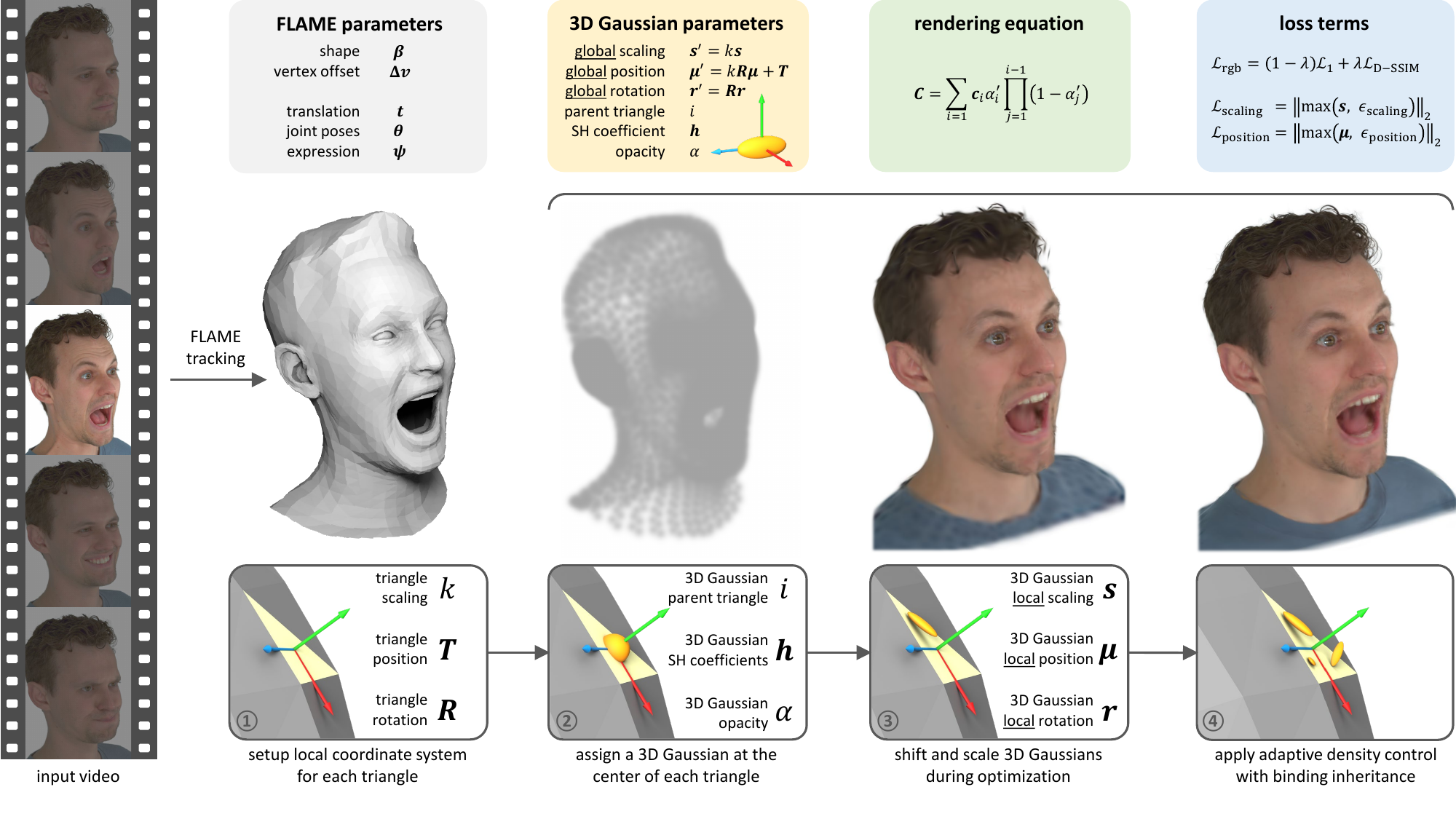}
  \caption{\textbf{Overview.} Our method binds 3D Gaussian splats to a FLAME \cite{FLAME:SiggraphAsia2017} mesh locally. We take the tracked mesh for each frame and transform the splats from local to global space before rendering them with 3D Gaussian Splatting \cite{kerbl20233d}. We optimize the splats in the local space by minimizing color loss on the rendering. We add and remove splats adaptively with their binding relation to triangles inherited so that all splats remain rigged throughout the optimization procedure. Further, we regularize the position and scaling of 3D Gaussian splats to suppress artifacts during animation.}
  \label{fig:pipeline}
\end{figure*}

As shown in \cref{fig:pipeline}, the input to our method is a multi-view video recording of a human head. For each time step, we use a photometric head tracker based on \cite{thies2016face2face} to fit FLAME \cite{FLAME:SiggraphAsia2017} parameters with multi-view observations and known camera parameters. The FLAME meshes have vertices at varied positions but share the same topology. Therefore, we can build a consistent connection between triangles of the mesh and 3D Gaussian splats (\cref{sec:rigging}). The splats are rendered into images via a differentiable tile rasterizer \cite{kerbl20233d}. These images are then supervised by the ground-truth images towards learning a photorealistic human head avatar. As per static scenes, it is also necessary to densify and prune Gaussian splats for optimal quality, with a set of adaptive density control operations \cite{kerbl20233d}. To achieve this without breaking the connection between triangles and splats, we design a binding inheritance strategy (\cref{sec:binding_inheritance}) so that new Gaussian points remain rigged to the FLAME mesh. In addition to the color loss, we also find it crucial to regularize the local position and scaling of Gaussian splats to avoid quality degradation under novel expressions and poses (\cref{sec:regularization}).

\subsection{Preliminary}
3D Gaussian Splatting \cite{kerbl20233d} provides a solution to reconstruct a static scene with anisotropic 3D Gaussians given images and camera parameters. A scene is represented by a set of Gaussian splats, each defined by a covariance matrix $\Sigma$ centered at point (mean) $\bm{\mu}$:
\begin{equation}
    G(\bm{x}) = e^{-\frac{1}{2}(\bm{x} - \bm{\mu})^T \Sigma^{-1}(\bm{x} - \bm{\mu})}
\end{equation}
Note that covariance matrices have physical meaning only when they are semi-definite, which cannot be guaranteed for an optimization process with gradient descent. Therefore, \citet{kerbl20233d} first define a parametric ellipse with a scaling matrix $S$ and a rotation matrix $R$, then construct the covariance matrix by:
\begin{equation}
    \Sigma = R S S^T R^T.
\end{equation}
Practically, an ellipse is stored as a position vector $\bm{\mu} \in \mathbb{R}^3$, a scaling vector $\bm{s} \in \mathbb{R}^3$, and a quaternion $\bm{q} \in \mathbb{R}^4$. In our paper, we use $\bm{r} \in \mathbb{R}^{3\times3}$ to notate the corresponding rotation matrix to $\bm{q}$.

For rendering, the color $\bm{C}$ of a pixel is computed by blending all 3D Gaussians overlapping the pixel:
\begin{equation}
    \bm{C} = \sum_{i=1}{\bm{c}_i \alpha_i' \prod_{j=1}^{i-1}}(1-\alpha_j'),
\end{equation}
where $\bm{c}_i$ is the color of each point modeled by 3-degree spherical harmonics. The blending weight $\alpha'$ is given by evaluating the 2D projection of the 3D Gaussian multiplied by a per-point opacity $\alpha$. The Gaussian splats are sorted by depth before blending to respect visibility order.

\subsection{3D Gaussian Rigging}
\label{sec:rigging}
The key component of our method is how we build connections between the FLAME \cite{FLAME:SiggraphAsia2017} mesh and 3D Gaussian splats. Initially, we pair each triangle of the mesh with a 3D Gaussian and let the 3D Gaussian move with the triangle across time steps. In other words, the 3D Gaussian is static in the local space of its parent triangle but dynamic in the global (metric) space as the triangle moves. Given the vertices and edges of a triangle, we take the mean position $\bm{T}$ of the vertices as the origin of the local space. Then, we concatenate the direction vector of one of the edges, the normal vector of the triangle, and their cross product as column vectors to form a rotation matrix $\bm{R}$, which describes the orientation of the triangle in the global space. We also compute a scalar $k$ by the mean length of one of the edges and its perpendicular to describe the triangle scaling.

For the paired 3D Gaussian of a triangle, we define its location $\bm{\mu}$, rotation $\bm{r}$, and anisotropic scaling $\bm{s}$ all in the local space. We initialize the location $\bm{\mu}$ at the local origin, the rotation $\bm{r}$ as an identity rotation matrix, and the scaling $\bm{s}$ as a unit vector. At rendering time, we convert these properties into the global space by:
\begin{align}
    \bm{r}' &= \bm{R} \bm{r}, \\
    \bm{\mu}' &= k \bm{R} \bm{\mu} + \bm{T}, \label{eq:global_position} \\
    \bm{s}' &= k \bm{s} \label{eq:global_scaling}.
\end{align}
We incorporate triangle scaling in \cref{eq:global_position,eq:global_scaling} so that the local position and scaling of a 3D Gaussian are defined relative to the absolute scale of a triangle. This enables an adaptive step size in the metric space with a constant learning rate for parameters defined in the local space. For example, a 3D Gaussian paired with a smaller triangle will move slower in an iteration step than those paired with large triangles. This also makes interpreting the parameters regarding the distance from the triangle center easier.

\subsection{Binding Inheritance}
\label{sec:binding_inheritance}
Only having the same numbers of Gaussian splats as the triangles is insufficient to capture details. For instance, representing a curved hair strand requires multiple splats, while a triangle on the scalp may intersect with several strands. Therefore, we also need the adaptive density control strategy \cite{kerbl20233d}, which adds and removes splats based on the view-space positional gradient and the opacity of each Gaussian. 

For each 3D Gaussian with a large view-space positional gradient, we split it into two smaller ones if it is large or clone it if it is small. We conduct this in the local space and ensure a newly created Gaussian is close to the old one that triggers this densification operation. Then, it is reasonable to bind a new 3D Gaussian to the same triangle as the old one because it was created to enhance the fidelity of the local region. Therefore, each 3D Gaussian must carry one more parameter, the index of its parent triangle, to enable binding inheritance during densification.

Besides densification, we also use the pruning operation as a part of the adaptive density control strategy \cite{kerbl20233d}. It periodically resets the opacity of all splats close to zero and removes points with opacity below a threshold. This technique is effective in suppressing floating artifacts, however, such pruning can also cause problems in a dynamic scene. For instance, regions of the face that are often occluded (such as eyeball triangles), can be overly sensitive to this pruning strategy, and often end up with few or no attached Gaussians. To prevent this, we keep track of the number of splats attached to each triangle, and ensure that every triangle always has at least one splat attached.

\subsection{Optimization and Regularization}
\label{sec:regularization}
We supervise the rendered images with a combination of $\mathcal{L}_1$ term and a D-SSIM term following \cite{kerbl20233d}:
\begin{equation}
    \mathcal{L}_{\text{rgb}} = (1-\lambda) \mathcal{L}_1 + \lambda \mathcal{L}_{\text{D-SSIM}},
\end{equation}
with $\lambda=0.2$. This already results in good re-rendering quality without additional supervision, such as depth or silhouette supervision, thanks to the powerful tile rasterizer \cite{kerbl20233d}. We found however, that if we try to animate these splats via FLAME \cite{FLAME:SiggraphAsia2017} to novel expressions and poses, large spike- and blob-like artifacts appear wildly throughout the scene. This is due to a poor alignment between the Gaussian splats and the triangles. 

\textbf{Position loss with threshold.}
A basic hypothesis behind 3D Gaussian rigging is that the Gaussian splats should roughly match the underlying mesh. They should also match their locations; for instance a Gaussian representing a spot on the nose should not be rigged to a triangle on the cheek. Although our splats are initialized at triangle centers, and new splats are added nearby these existing ones, it is not guaranteed that the primitives remain close to their parent triangle after the optimization. To address this, we regularize the local \textit{position} of each Gaussian by:
\begin{equation}
    \mathcal{L}_{\text{position}} = \| \max \left( \bm{\mu}, \epsilon_{\text{position}} \right) \|_2,
\end{equation}
where $\epsilon_{\text{position}}=1$ is a threshold that tolerates small errors within the scaling of its parent triangle. 

\textbf{Scaling loss with threshold.}
Aside from position, the scaling of 3D Gaussians is even more essential for the visual quality during animation. Specifically, if a 3D Gaussian is large in comparison to its parent triangle, small rotations of the triangle -- barely noticeable at the scale of the triangle -- will be magnified by the scale of the 3D Gaussian, resulting in unpleasant jittering artifacts. To mitigate this, we also regularize the local \textit{scale} of each 3D Gaussian by:
\begin{equation}
    \mathcal{L}_{\text{scaling}} = \| \max \left( \bm{s}, \epsilon_{\text{scaling}} \right) \|_2,
    \label{eq:scaling_loss}
\end{equation}
where $\epsilon_{\text{scaling}}=0.6$ is a threshold that disables this loss term when the scale of a Gaussian less than $0.6\times$ the scale of its parent triangle.
This $\epsilon_{\text{scaling}}$-tolerance is indispensable; without it, the Gaussian splats shrink excessively, causing rendering speeds to deteriorate, as camera rays need to hit more splats before zero transmittance is reached.

Our final loss function is thus:
\begin{equation}
    \mathcal{L} = \mathcal{L}_{\text{rgb}} + \lambda_{\text{position}} \mathcal{L}_{\text{position}} + \lambda_{\text{scaling}} \mathcal{L}_{\text{scaling}},
\end{equation}
where $\lambda_{\text{position}}=0.01$ and $\lambda_{\text{scaling}}=1$. Note that we only apply $\mathcal{L}_{\text{position}}$ and $\mathcal{L}_{\text{scaling}}$ to visible splats. Thereby, we only regularize points when the color loss $\mathcal{L}_{\text{rgb}}$ is present. This helps to maintain the learned structure of often-occluded regions such as teeth and eyeballs.

\textbf{Implementation details.}
We use Adam \cite{kingma2014adam} for parameter optimization (the same hyperparameter values are used across all subjects). We set the learning rate to 5e-3 for the position and 1.7e-2 for the scaling of 3D Gaussians and keep the same learning rates as 3D Gaussian Splatting \cite{kerbl20233d} for the rest of the parameters. Alongside the Gaussian splat parameters, we also finetune the translation, joint rotation, and expression parameters of FLAME \cite{FLAME:SiggraphAsia2017} for each timestep, using learning rates 1e-6, 1e-5, and 1e-3, respectively. We train for 600,000 iterations, and exponentially decay the learning rate for the splat positions until the final iteration, where it reaches 0.01$\times$ the initial value. We enable adaptive density control with binding inheritance every 2,000 iterations, from iteration 10,000 until the end. Every 60,000 iterations, we reset the Gaussians' opacities. We use a photo-metric head tracker to obtain the FLAME parameters, including shape $\bm{\beta}$, translation $\bm{t}$, pose $\bm{\theta}$, expression $\bm{\psi}$, and vertex offset $\Delta \bm{v}$ in the canonical space. 

%% file: sec/4_experiments.tex
\section{Experiments}
\label{sec:exp}

\begin{figure*}[h]
  \centering
  \includegraphics[width=1.\textwidth ]{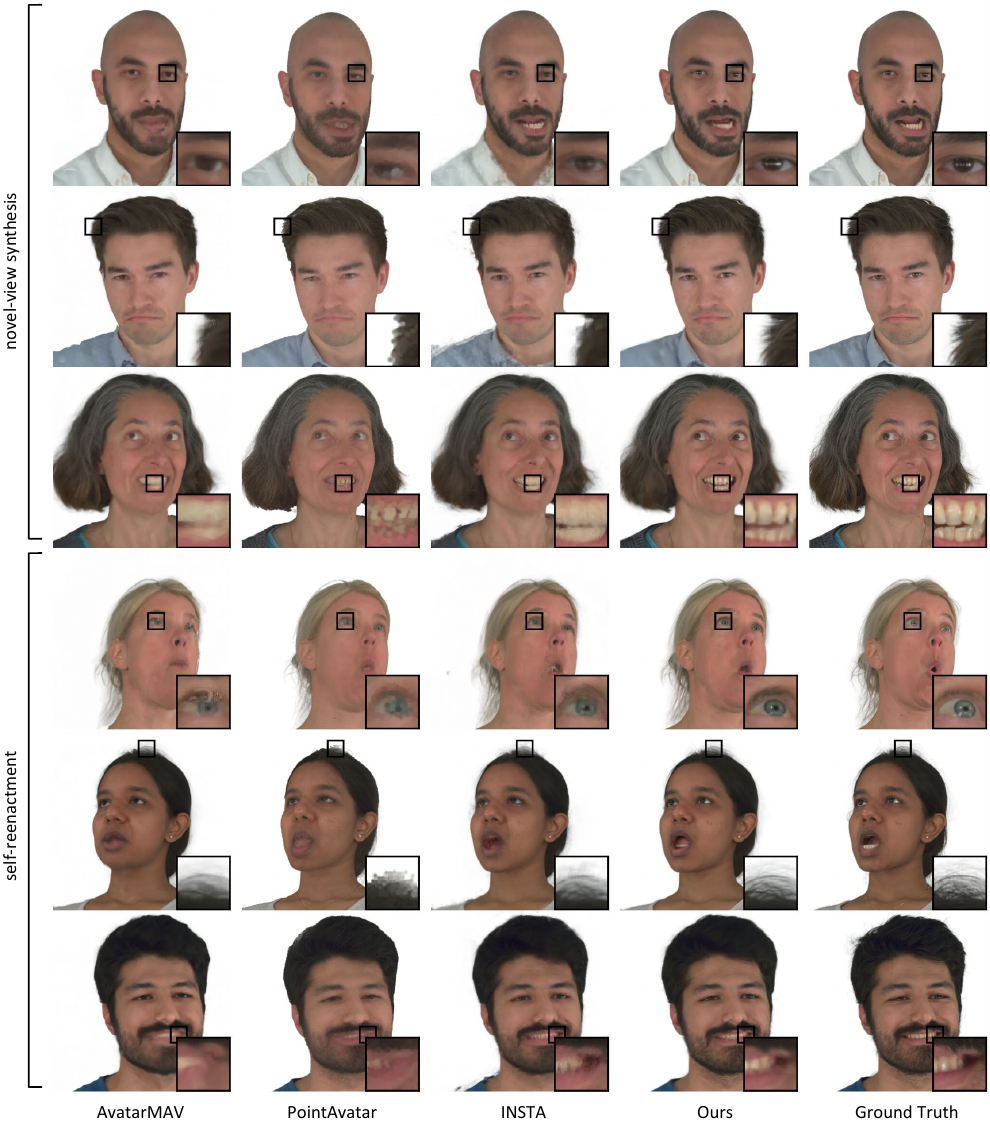}
  \caption{Qualitative comparison on novel-view synthesis and self-reenactment of head avatars. Our method outperforms state-of-the-art methods by producing significantly sharper rendering outputs. We obtain precise reconstruction of details such as reflective light on eyes, hair strands, teeth, etc. Our results for self-reenactment show more accurate expressions compare to baselines.}
  \label{fig:novel_view_expr}
\end{figure*}

\begin{figure*}[h]
  \centering
  \includegraphics[width=1\textwidth ]{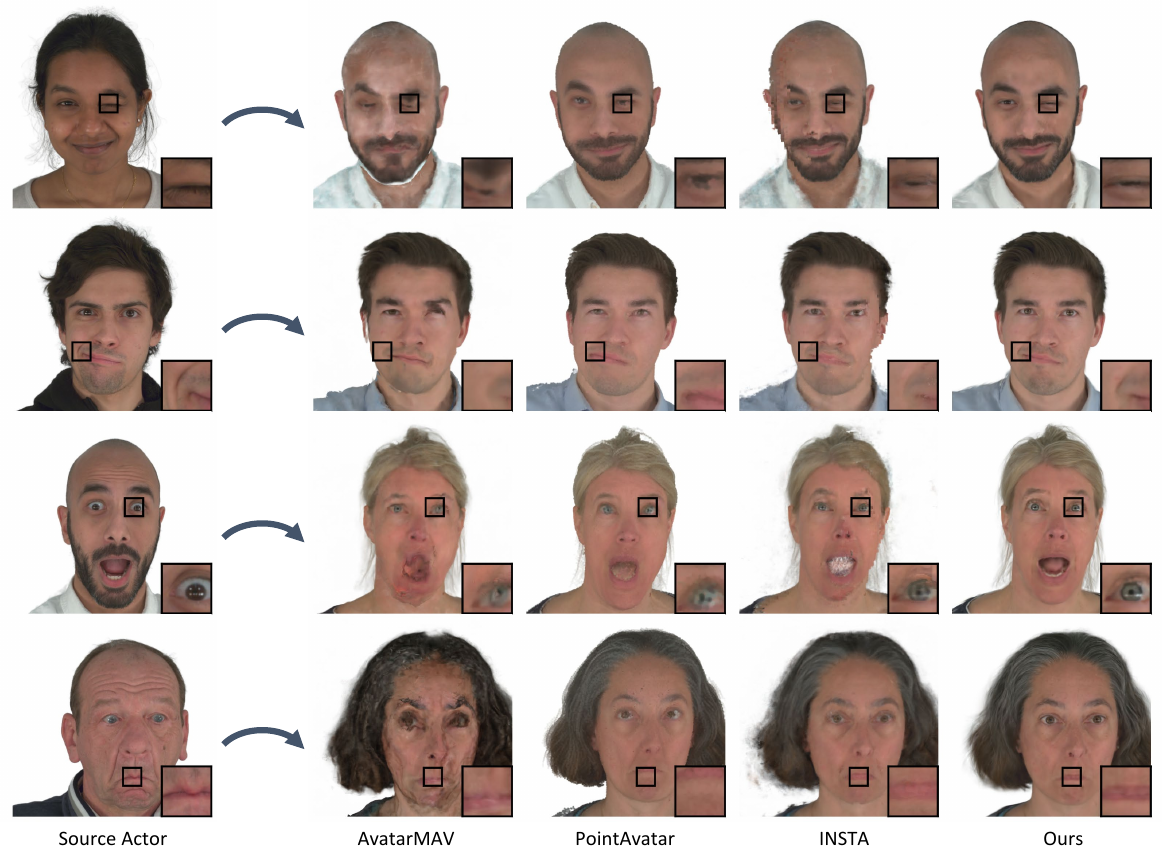}
  \caption{Cross-identity reenactment of head avatars. We use the tracked FLAME expression and pose parameters of source actors to drive the reconstructed avatars. Our method produces high-quality rendering and transfers expressions vividly, while baseline methods suffer from artifacts and generalize poorly to novel expressions.}
  \label{fig:reenactment}
\end{figure*}

\subsection{Setup}

\textbf{Settings.} We evaluate the quality of head avatars with three settings: 1) \textit{novel-view synthesis}: driving an avatar with head poses and expressions from training sequences and rendering from a held-out viewpoint. 2) \textit{self-reenactment}: driving an avatar with unseen poses and expressions from a held-out sequence of the same subject and rendering all 16 camera views. 3) \textit{cross-identity reenactment}: animating an avatar with poses and expressions from another subject.

\noindent\textbf{Dataset.} We conduct experiments on video recordings of 9 subjects from the NeRSemble \cite{kirschstein2023nersemble} dataset. All recordings contain 16 views covering the front and sides of a subject. For each subject, we take 11 video sequences and downsample images to a resolution of $802 \times 550$. 
Participants were instructed to perform specific expressions or emotions in 10 sequences and freely perform in the last one.
To conduct quantitative evaluation for each subject, we train our method and three baselines (INSTA~\cite{zielonka2023instant}, PointAvatar~\cite{zheng2023pointavatar}, and AvatarMAV~\cite{xu2023avatarmav}) using 9 out of the 10 prescribed sequences and 15 out of 16 available cameras. We use the 11\ts{th} (free-performance) sequence to visually assess cross-identity reenactment capabilities.
Please refer to the supplementary material for details of the dataset and baselines.

\begin{table}[t]
\small
\renewcommand{\arraystretch}{1}
\setlength{\tabcolsep}{1.1pt}
\begin{tabular}{@{}l|ccc|ccc@{}}
\toprule
                   & \multicolumn{3}{c|}{Novel-View Synthesis}                                                    & \multicolumn{3}{c}{\cellcolor[HTML]{FFFFFF}Self-Reenactment}                                 \\ \cmidrule(l){2-7} 
\multirow{-2}{*}{} & PSNR↑                        & SSIM↑                         & LPIPS↓                        & PSNR↑                        & SSIM↑                         & LPIPS↓                        \\ \midrule
AvatarMAV \cite{xu2023avatarmav}          & \cellcolor[HTML]{FFF2CC}29.5 & \cellcolor[HTML]{FFF2CC}0.913 & 0.152                         & 24.3                         & 0.887                         & 0.168                         \\
PointAvatar \cite{zheng2023pointavatar}        & 25.8                         & 0.893                         & \cellcolor[HTML]{FFF2CC}0.097 & 23.4                         & 0.884                         & \cellcolor[HTML]{FFF2CC}0.102 \\
INSTA \cite{zielonka2023instant}             & 26.7                         & 0.899                         & 0.122                         & \cellcolor[HTML]{E2EFDA}26.3 & \cellcolor[HTML]{FFF2CC}0.906 & 0.110                         \\
Ours               & \cellcolor[HTML]{E2EFDA}31.6 & \cellcolor[HTML]{E2EFDA}0.938 & \cellcolor[HTML]{E2EFDA}0.065 & \cellcolor[HTML]{FFF2CC}26.0 & \cellcolor[HTML]{E2EFDA}0.910 & \cellcolor[HTML]{E2EFDA}0.076 \\ \bottomrule
\end{tabular}
\caption{Quantitative comparison with state-of-the-art methods. \colorbox{green}{Green} indicates the best and \colorbox{yellow}{yellow} indicates the second.}
\label{tab:sota}
\end{table}

\subsection{Head Avatar Reconstruction and Animation}

We evaluate the reconstruction quality of avatars by novel-view synthesis and animation fidelity by self-reenactment. \cref{fig:novel_view_expr} shows qualitative comparisons. For novel-view synthesis, all methods produce plausible rendering results. A close inspection of PointAvatar's \cite{zheng2023pointavatar} results show dotted artifacts, owing its fixed point size. In our case, the anisotropic scaling of 3D Gaussians alleviates this issue.

INSTA \cite{zielonka2023instant} shows clean results for the face, however regions around the neck and shoulder can be noisy, as the tracked FLAME meshes are often misaligned in those regions. This causes issues for INSTA, as its warping process is based on the nearest triangle. In the case of our method, each Gaussian splat is rigged to a consistent triangle, regardless of the pose or expression. When the tracked mesh is inaccurate, the positional gradient to a Gaussian splat can consistently back-propagate to the same triangle. This enables misalignments due to incorrect FLAME tracking to be corrected while the 3D Gaussians are being optimized.

AvatarMAV \cite{xu2023avatarmav} shows comparable quality to other methods in novel-view synthesis but struggles with novel-expression synthesis. This is because it only uses the expression vector of a 3DMM as conditioning. Since the control from the expression vectors to the deformation bases must be learned, it struggles to reproduce expressions that are far from the training distribution. Similar conclusions can be drawn from the quantitative comparison shown in \cref{tab:sota}. Our approach outperforms others by a large margin regarding metrics for novel-view synthesis. Our method also stands out in self-reenactment, with significantly lower perceptual differences in terms of LPIPS \cite{zhang2018unreasonable}.
Note that self-reenactment is based on tracked FLAME meshes that may not perfectly align with the target images, thus bringing disadvantages to our results with more visual details regarding pixel-wise metrics such as PSNR.

For a real-world test of avatar animation, we conduct experiments on cross-identity reenactment in \cref{fig:reenactment}. Our avatars accurately reproduce eye blinks and mouth movements from source actors showing lively, complex dynamics such as wrinkles. INSTA \cite{zielonka2023instant} suffers from aliasing artifacts when the avatars move beyond the occupancy grid of I-NGP \cite{muller2022instant} optimized for training sequences. The movement of results from PointAvatar \cite{zheng2023pointavatar} is not precise because its deformation space is not guaranteed to be consistent with FLAME. AvatarMAV \cite{xu2023avatarmav} exhibits large degradations in reenactment due to a lack of deformation priors.

\subsection{Ablation Study}

\begin{figure}[t]
  \centering
  \includegraphics[width=1\linewidth]{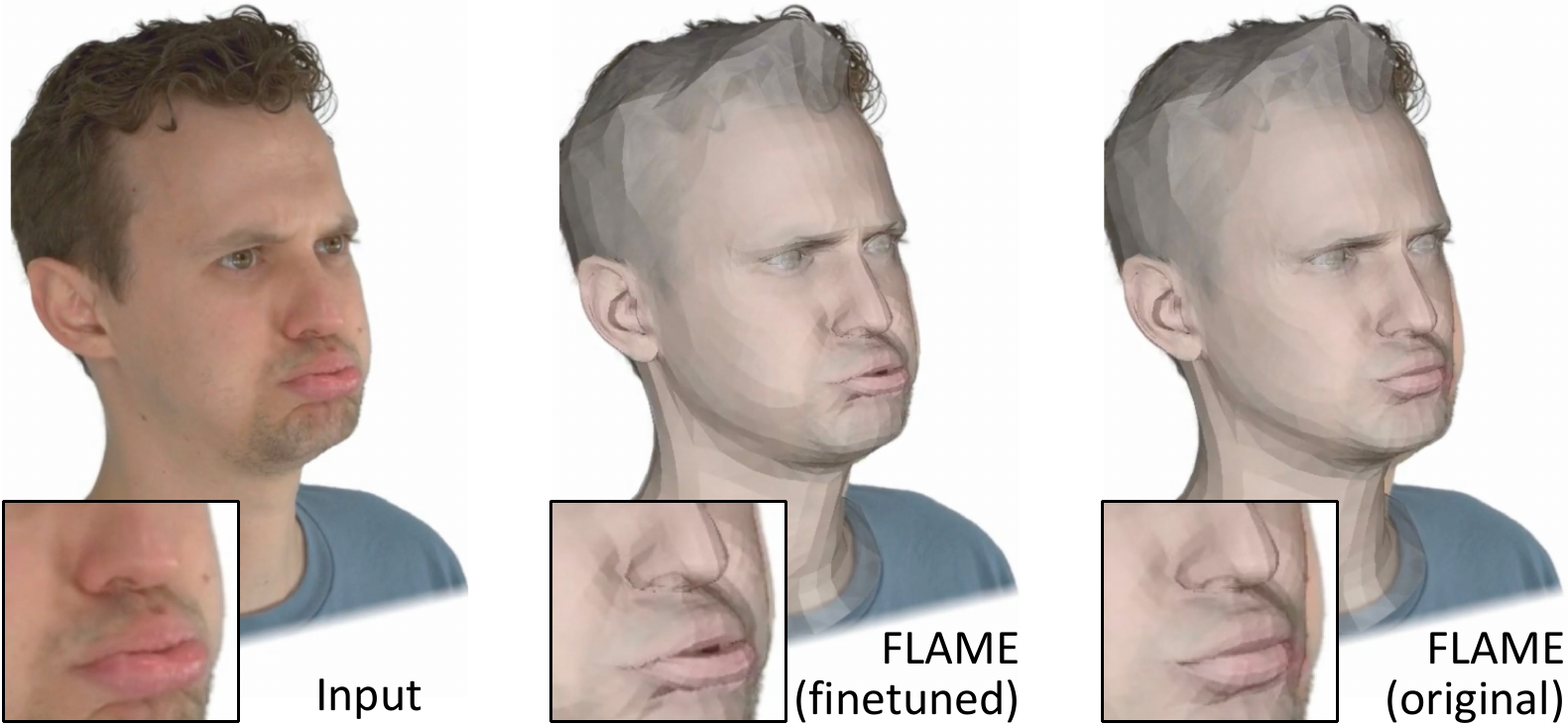}
   \caption{Fine-tuning FLAME parameters leads to better alignment of the mesh to the input image. In the example above, the movements of cheeks and lips are better captured after fine-tuning.}
   \label{fig:flame_finetune}
\end{figure}

To validate the effectiveness of our method components, we deactivate each of them and report results in \cref{tab:ablation}.

\textbf{Adaptive density control with binding inheritance.}
Without binding inheritance, we lose the ability to add and remove Gaussian splats besides the initial ones. With a limited number of splats, the scaling of each splat increases to occupy the same space, causing blurry renderings, leading to huge fidelity loss, as shown in the second row of \cref{tab:ablation}.

\begin{figure}[tb]
  \centering
  \includegraphics[width=1\linewidth ]{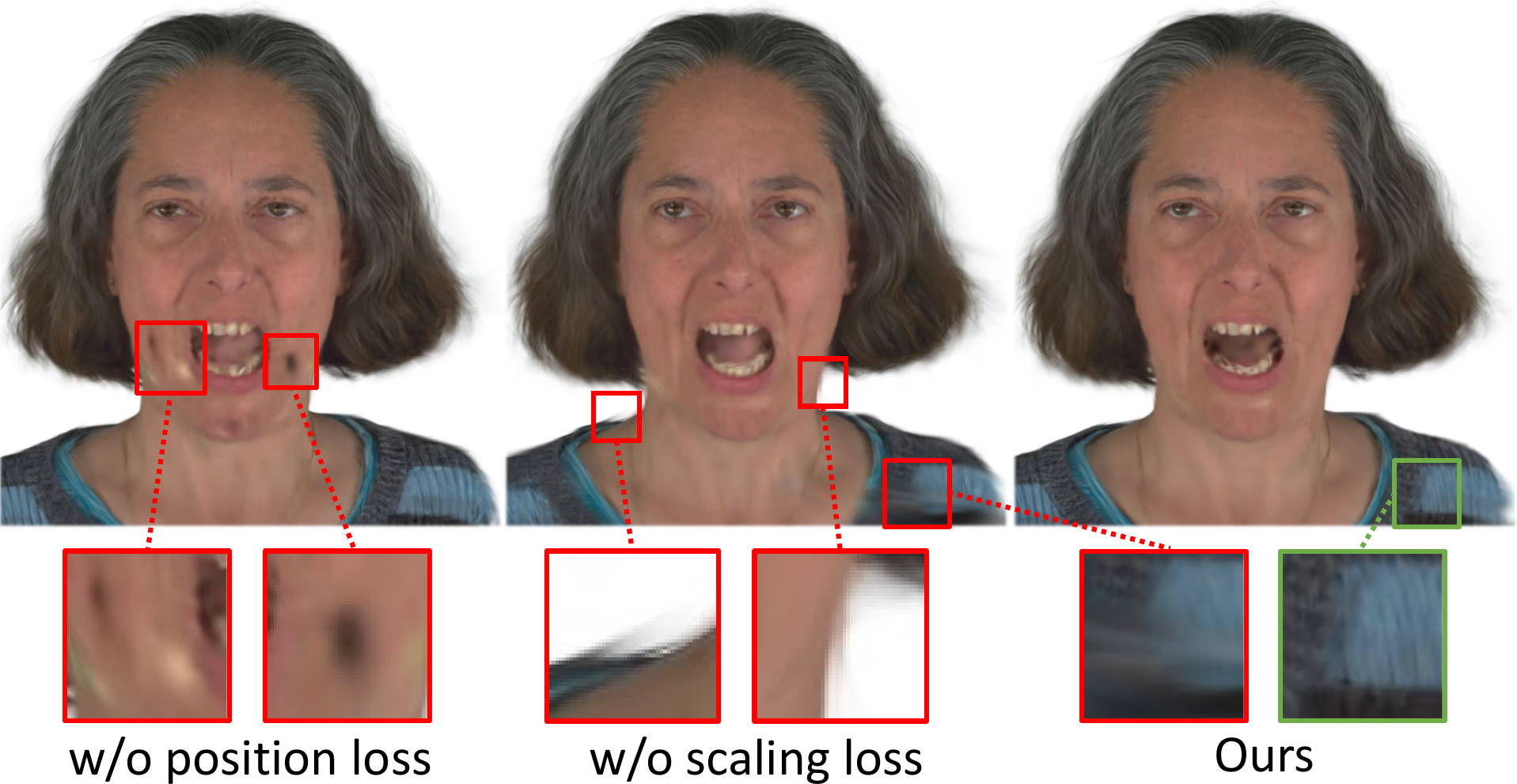}
  \caption{ The position loss and the scaling loss helps prevent artifacts during animation with novel expressions and poses.}
  \label{fig:position_scale_loss}
\end{figure}

\textbf{Regularization on the scaling of Gaussian splats.}
Without the scaling loss, artifacts show up as spikes (the middle column in \cref{fig:position_scale_loss}), and slightly worsen the image quality (the third row in \cref{tab:ablation}). But when we use the scaling loss without an error tolerance, all metrics drastically deteriorate (the fourth row in \cref{tab:ablation}) because all the splats are regularized to be infinitely small, making it hard to construct opaque surfaces.

\textbf{Regularization on the position of Gaussian splats.}
When disabling the threshold for the position loss, we get slightly worse metrics (the sixth row in \cref{tab:ablation}). But when turning off the position loss, the metrics on novel-view synthesis become the best in the table (the fifth row in \cref{tab:ablation}). This is reasonable because without constraints to the mesh Gaussian splats can freely move in the space to achieve minimal re-rendering error. However, since the distribution of Gaussian splats are overfitted to training frames, the avatar shows artifacts such as cracks and fly-around blobs (the left column in \cref{fig:position_scale_loss}) in novel expressions and poses. Therefore, the position loss is necessary to enforce a conservative deviation of splats from the mesh surface so that we can animate the reconstructed avatar with unseen facial motion.

\textbf{FLAME fine-tuning.}
Thanks to our consistent binding between Gaussian splats and triangles, we can effectively fine-tune FLAME parameters while optimizing Gaussian splats (\cref{fig:flame_finetune}). As shown in the last row in \cref{tab:ablation}, turning off the fine-tuning negatively affects image quality for novel-view synthesis. For self-reenactment, fine-tuning FLAME parameters also leads to lower perceptual error.

\begin{table}[]
\small
\renewcommand{\arraystretch}{1.0}
\setlength{\tabcolsep}{1pt}
\begin{tabular}{@{}l|ccc|ccc@{}}
\toprule
\multicolumn{1}{c|}{}                   & \multicolumn{3}{c|}{Novel-View}                                                              & \multicolumn{3}{c}{\cellcolor[HTML]{FFFFFF}Self-Reenactment}                                 \\ \cmidrule(l){2-7} 
\multicolumn{1}{c|}{\multirow{-2}{*}{}} & \multicolumn{1}{c}{PSNR↑}    & \multicolumn{1}{c}{SSIM↑}     & \multicolumn{1}{c|}{LPIPS↓}   & \multicolumn{1}{c}{PSNR↑}    & \multicolumn{1}{c}{SSIM↑}     & \multicolumn{1}{c}{LPIPS↓}    \\ \midrule
Ours                                    & \cellcolor[HTML]{FFF2CC}28.8 & \cellcolor[HTML]{FFF2CC}0.883 & \cellcolor[HTML]{FFF2CC}0.098 & \cellcolor[HTML]{FFF2CC}25.1 & 0.853                         & \cellcolor[HTML]{FFF2CC}0.101 \\
w/o   ADC                     & 26.8                         & 0.854                         & 0.206                         & \cellcolor[HTML]{FFF2CC}25.1 & \cellcolor[HTML]{FFF2CC}0.860 & 0.183                         \\
w/o   $\mathcal{L}_{\text{scaling}}$    & 28.0                         & 0.877                         & 0.114                         & 24.9                         & 0.852                         & 0.109                         \\
w/o   $\epsilon_{\text{scaling}}$       & 25.0                         & 0.833                         & 0.195                         & 24.1                         & 0.843                         & 0.176                         \\
w/o   $\mathcal{L}_{\text{position}}$   & \cellcolor[HTML]{E2EFDA}29.7 & \cellcolor[HTML]{E2EFDA}0.894 & \cellcolor[HTML]{E2EFDA}0.091 & 24.9                         & 0.851                         & \cellcolor[HTML]{E2EFDA}0.096 \\
w/o   $\epsilon_{\text{position}}$      & 28.7                         & 0.882                         & 0.105                         & 25.0                         & 0.855                         & 0.106                         \\
w/o   FLAME ft.                         & 26.1                         & 0.855                         & 0.131                         & \cellcolor[HTML]{E2EFDA}25.5 & \cellcolor[HTML]{E2EFDA}0.862 & 0.124                         \\ \bottomrule
\end{tabular}
\caption{Ablation study on subject \#304. \colorbox{green}{Green} indicates the best and \colorbox{yellow}{yellow} indicates the second. ``ADC" refers to Adaptive Density Control with binding inheritance. `FLAME ft.' refers to FLAME parameter fine-tuning.}
\label{tab:ablation}
\end{table}

%% file: sec/5_conclusion.tex
\section{Limitations and Potential Negative Impacts}
Our approach, based on 3D Gaussian splats, directly captures the radiance field without decoupling material and lighting. Hence, relighting the avatar is not feasible within our current approach. 
Additionally, we lack control over areas that are not modeled by FLAME such as hair and other accessories. Here, we believe that modeling these parts directly would provide an interesting future research avenue; for instance, recent methods show promising result in reconstructing hair \cite{rosu2022neural, wang2023neuwigs}.

The creation of photorealistic avatars may introduce a wide spectrum of malevolent uses.
Privacy violations are a significant concern, with potential for unauthorized manipulation of people's likeness. Moreover the creation of deceptive content, like deepfake videos, can lead to misinformation, defamation and harm to reputations. Identity theft is an additional risk, as avatars may be exploited for impersonation and fraudulent activities. We strongly condemn unauthorized and malevolent uses of this technology, emphasizing the need for ethical considerations in all applications utilizing our method.

\section{Conclusion}
\OURS{} is a novel approach that creates photorealistic avatars from video sequences. It features a dynamic 3D representation based on 3D Gaussian splats that are rigged to a parametric morphable face model. This enables flexible control and precise attribute transfer of a dynamic, photorealistic avatar. The set of Gaussian splats can deviate from the mesh surface to compensate the absence or inaccuracy of the morphable model, exhibiting astonishing ability to model fine details on human heads. 
Our approach outperforms state-of-the-art methods on image quality and expression accuracy for a large margin, indicating a large potential for more applications in related domains.

\section*{Acknowledgements}
This work was supported by Toyota Motor Europe and Woven by Toyota. This work was also supported by the ERC Starting Grant Scan2CAD (804724). We thank Justus Thies, Andrei Burov, and Lei Li for constructive discussions and Angela Dai for the video voice-over.

%% file: sec/x_suppl.tex
\appendix

\section{FLAME Tracking}
\label{sec:tracking}

For FLAME \cite{FLAME:SiggraphAsia2017} tracking, we optimize for per-frame parameters (translation $\bm{t}_i$, joint poses $\bm{\theta}_i$, expression $\bm{\psi}_i$) and shared parameters (shape $\bm{\beta}$, vertex offset $\Delta \bm{v}$, and an albedo map $A$). Our optimization combines a landmark loss, a color loss, and regularization terms.

We use a state-of-the-art facial landmark detector \cite{zhou2023star} to obtain 68 facial landmarks in 300-W \cite{sagonas2016300} format. Among them, we exclude 17 facial contour landmarks to avoid inconsistency caused by occlusion. 
We use NVDiffRast \cite{laine2020diffrast} to render FLAME meshes and obtain gradients of vertex positions regarding the color loss by texel interpolation for the interior and anti-aliasing on the boundary. For regularization, we apply a Laplacian smoothness term on the vertex offset and temporal smoothness terms on the per-frame parameters.

We optimize all the parameters on the first time step of the video sequence until convergence, then optimize per-frame parameters for 50 iterations for each following time step with the previous one as initialization. Afterward, we conduct global optimization for 30 epochs by randomly sampling time steps to fine-tune all parameters.

\begin{figure}[h]
  \centering
  \includegraphics[width=1\linewidth ]{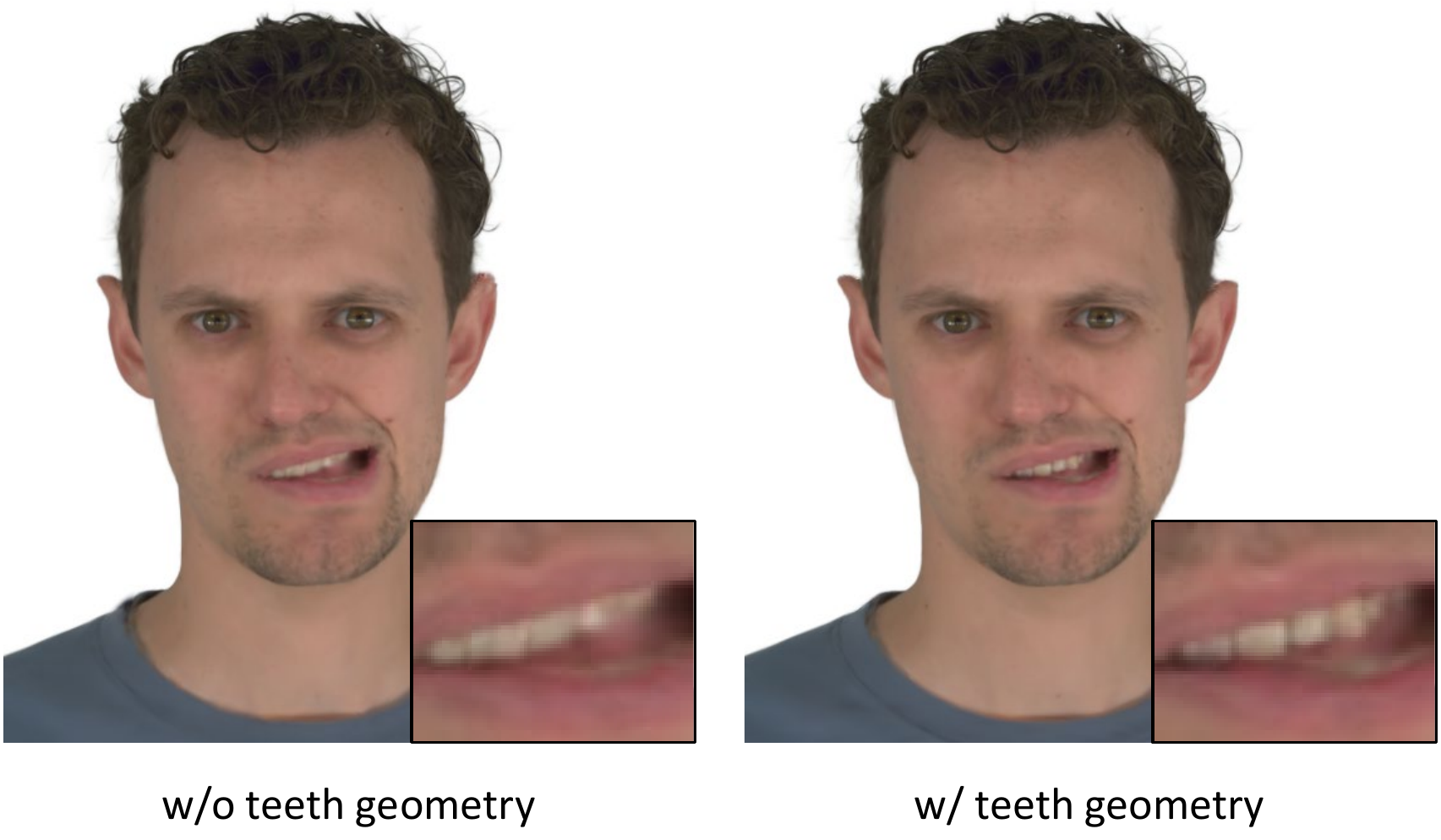}
  \caption{Adding triangles that move rigidly with the head and the jaw helps Gaussian splats to capture teeth details.}
  \label{fig:teeth}
\end{figure}

We use the 2023 version of FLAME \cite{FLAME:SiggraphAsia2017} for the revised eye regions. Furthermore, we manually add 168 triangles for teeth to the template mesh of FLAME and make the upper and lower teeth triangles rigid to the neck and jaw joints, respectively. This improves the fidelity of our avatar as shown in \cref{fig:teeth}.

\section{Dataset Division}
\label{sec:data_proc}

We use 11 sequences for each subject from the NeRSemble~\cite{kirschstein2023nersemble} dataset. \cref{tab:sequences} lists concrete sequence types and IDs for different settings. 
Among the emotion (EMO) and expression (EXP) sequences, we randomly hold out one for self-reenactment evaluation (\cref{tab:test-sequence}) and use the rest nine for training.

\begin{figure}[t]
  \centering
  \includegraphics[width=1\linewidth ]{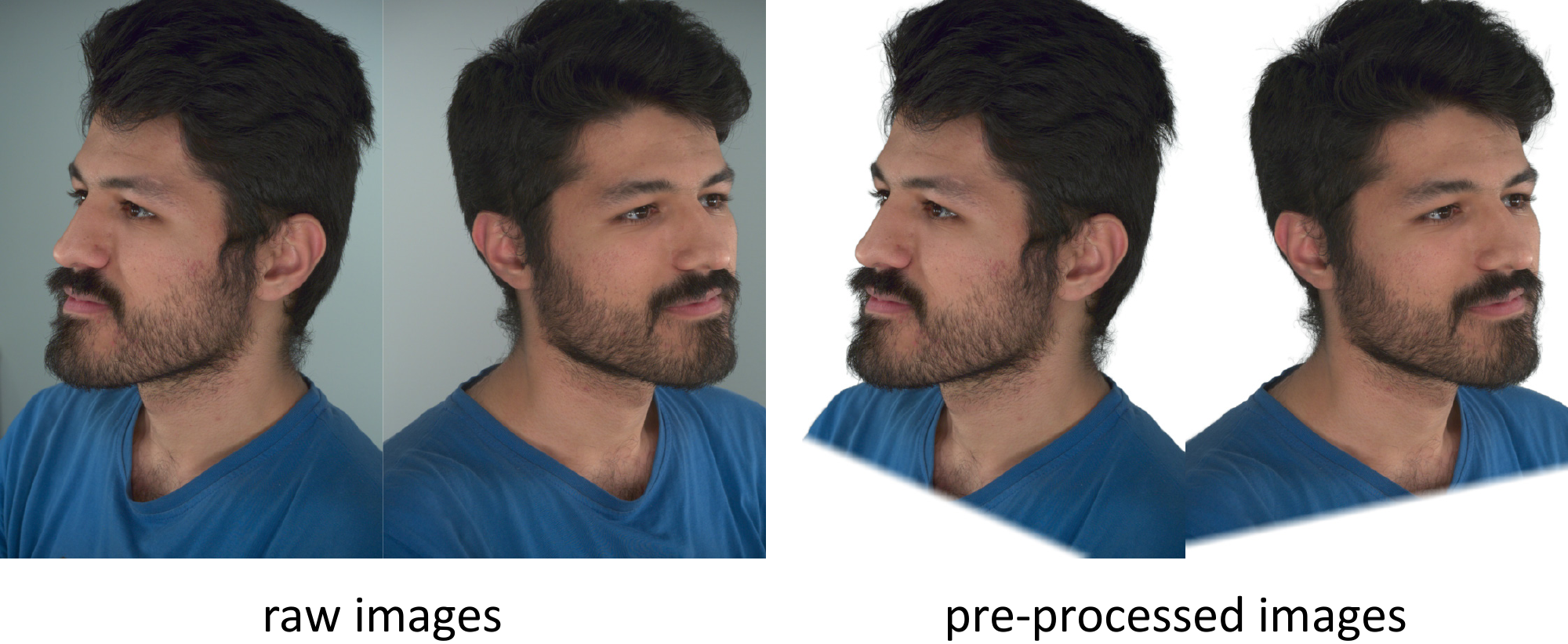}
  \caption{We remove the background and pixels below the shoulder to focus on the head region.}
  \label{fig:preprocess}
\end{figure}

To simplify the pipeline for Gaussian splat optimization, we remove the background of raw images with Background Matting V2 \cite{lin2021real}. Additionally, we fit a line across the bottom vertices of each tracked FLAME mesh and project the line to each viewpoint to remove the pixels below. We show an example of pre-processing results in \cref{fig:preprocess}.

\begin{table}[t]
\centering
\setlength{\tabcolsep}{5pt}
\scriptsize
\begin{tabular}{@{}l|cccccccccc|c@{}}
\toprule
Setting          & \multicolumn{10}{c|}{\begin{tabular}[c]{@{}c@{}}Novel View Synthesis\\ \& Self-reenactment\end{tabular}} & \multicolumn{1}{l}{\begin{tabular}[c]{@{}l@{}}Cross-identity\\ Reenactment\end{tabular}} \\ \midrule
Sequence Type & \multicolumn{4}{c|}{EMO}                          & \multicolumn{6}{c|}{EXP}                & FREE                                                                                     \\ \midrule
Sequence ID   & 1      & 2      & 3      & \multicolumn{1}{c|}{4}      & 2      & 3      & 4     & 5     & 8     & 9     & -                                                                                        \\ \bottomrule
\end{tabular}
\caption{The types and IDs of sequences for different settings.}
\label{tab:sequences}
\end{table}

\begin{table}[t]
\centering
\scriptsize
\setlength{\tabcolsep}{0.7pt}
\begin{tabular}{@{}l|ccccccccc@{}}
\toprule
Subject ID    & 074   & 104   & 218   & 253   & 264   & 302   & 304   & 306   & 460   \\ \midrule
Test Sequence & EMO-4 & EXP-2 & EXP-9 & EMO-4 & EXP-9 & EMO-2 & EXP-2 & EXP-2 & EMO-3 \\ \bottomrule
\end{tabular}
\caption{The held-out sequence of each subject for self-reenactment evaluation.}
\label{tab:test-sequence}
\end{table}

\begin{figure*}[t]
\centering
\begin{subfigure}{0.49\linewidth}
    \includegraphics[width=1\textwidth]{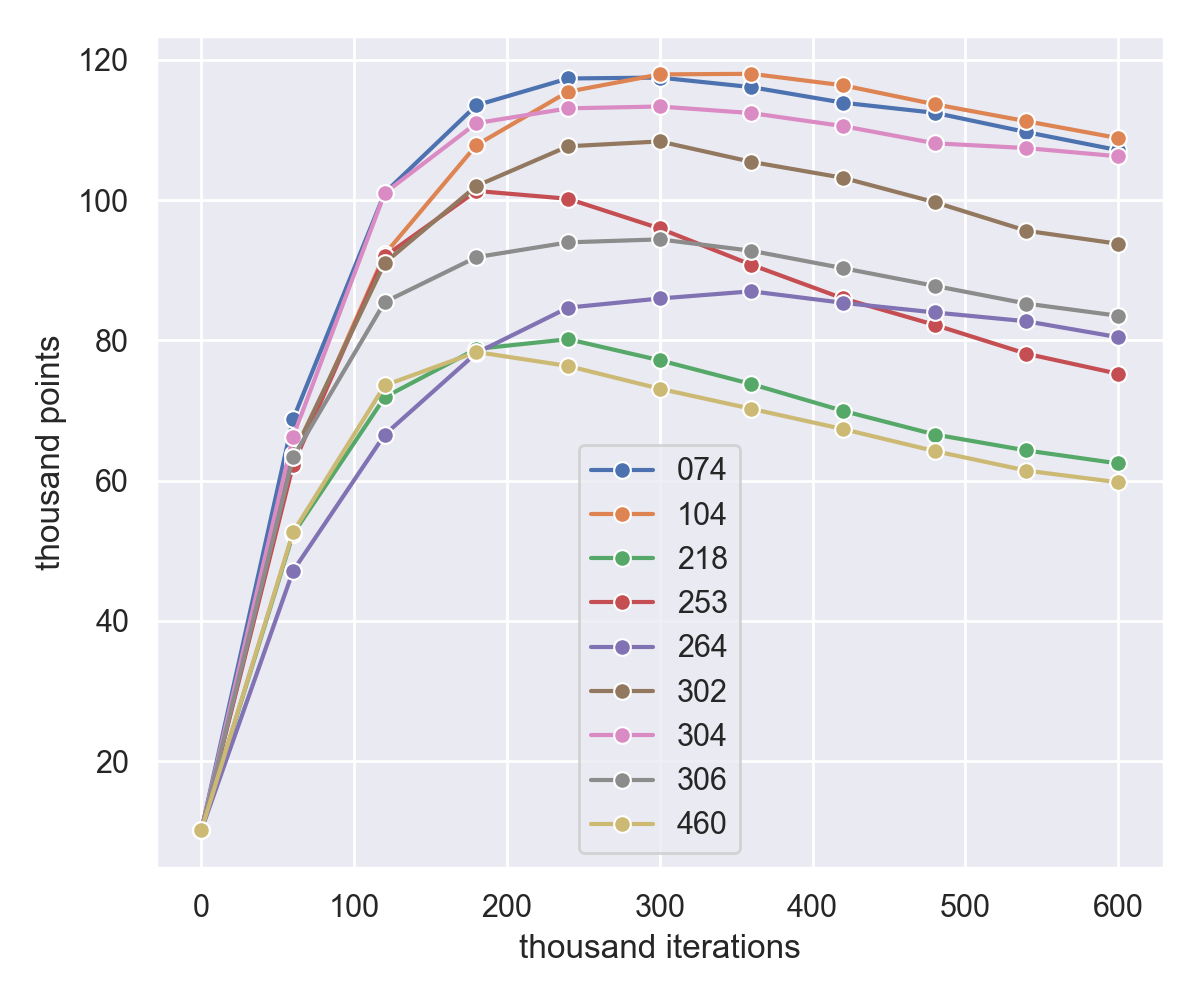}
    \caption{The number of 3D Gaussians throughout the optimization process.}
    \label{fig:num_points}
\end{subfigure}
\hfill
\begin{subfigure}{0.49\linewidth}
    \includegraphics[width=1\textwidth]{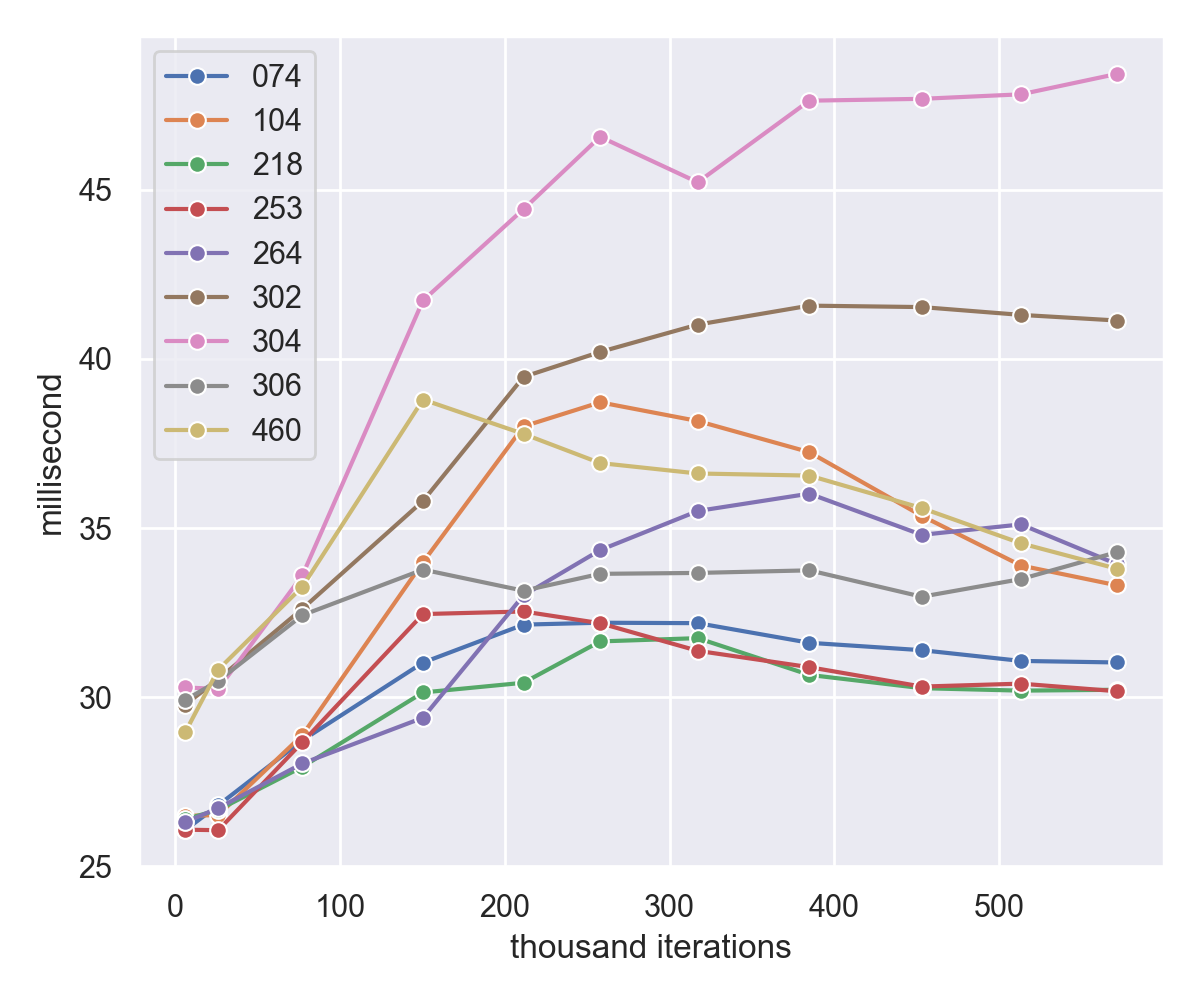}
    \caption{The run-time of an optimization iteration.}
    \label{fig:iter_time}
\end{subfigure}
\caption{The number of 3D Gaussians increases by a factor of around 10 from its starting point for all subjects. After this, the number of 3D Gaussians stops growing. Despite this growth in Gaussians, the run time of each training iteration at most only doubles. Each curve corresponds to a different subject.}
\label{fig:pts_runtime}
\end{figure*}

\section{Computation Efficiency}
Our method binds 3D Gaussians to triangles in an efficient way, maintaining high rendering and optimization speed. Given that 3D Gaussians are actively added and pruned during optimization, the running speed of the program also changes. 

\textbf{Efficiency during optimization.}
We show the evolution of the number of Gaussians and the run-time of an iteration in \cref{fig:pts_runtime}. According to \cref{fig:num_points}, the number of Gaussians grows from 10,144 (that is, the number of triangles in our modified FLAME mesh) to around 100,000 (on average). After this point is reached however, the number of Gaussians no longer increases. Thanks to this, longer training times do not mean ever-increasing memory requirements. In fact, our model can fit and be trained on an NVIDIA RTX 2080 Ti Graphics card with 12 gigabytes of VRAM. Moreover, while the number of Gaussians grows by as much as 1000\% during training, the run-time of each optimization iteration increases by less than 100\% (\cref{fig:iter_time}) at this peak. This validates the efficiency of the differentiable tile rasterizer \cite{kerbl20233d}, which sorts splats before blending and terminates ray marching once zero transmittance is reached. The threshold of our scaling loss (see Section 3 of the main paper) is crucial to this efficiency. Without it, rendering time would increase substantially, as the rasterizer would need to blend many more Gaussians before reaching zero transmittance.

\textbf{Efficiency during inference.}
Although our data are processed into a fixed resolution, the optimized model can be rendered in arbitrary resolutions. We show the average rendering FPS in variant resolutions to validate the efficiency of our method to suffice real-time applications.

\begin{table}[h]
\centering
\setlength{\tabcolsep}{4.7pt}
\scriptsize
\begin{tabular}{@{}c|ccccc@{}}
\toprule
Resolution & 401$\times$225 & 802$\times$550 & 1604$\times$1100 & 3208$\times$2200 & 6416$\times$4400 \\ \midrule
FPS        & 187     & 187     & 156       & 95        & 36        \\ \bottomrule
\end{tabular}
\caption{Rendering speed tested on subject \#306.}
\end{table}

\section{Baselines}
We compare our method with three state-of-the-art methods for head avatar creation.

INSTA \cite{zielonka2023instant} directly warps points according the nearest FLAME \cite{FLAME:SiggraphAsia2017} mesh triangle. It adds triangles to the mouth, and conditions radiance field queries in the mouth region on the expression code of FLAME to improve the quality of the mouth interior. The loss weight for the mouth region is 40$\times$ higher than other regions. It also applies a depth loss on the facial region.

PointAvatar \cite{zheng2023pointavatar} uses a point-based representation, which is closely related to 3D Gaussians.  It does not directly rely on the FLAME surface but uses its pose and expression parameters to condition a deformation field. During optimization, it applies a coarse-to-fine strategy to progressively increase the size of the point cloud and decrease the radius of each point. It also uses a post-processing operation to fill holes by applying erosion and dilation to rendered images.

AvatarMAV \cite{xu2023avatarmav} uses voxel grids for both a canonical radiance field and a set of bases of a motion field. It models deformation by blending the motion bases with the tracked expression vectors of a 3D morphable model \cite{paysan20093d}. We adapt this method to use our tracked FLAME poses and expressions to ensure fairness.